\documentclass[10pt,twocolumn,letterpaper]{article}

\usepackage{cvpr}
\usepackage{times}
\usepackage{epsfig}
\usepackage{graphicx}
\usepackage{amsmath}
\usepackage{amssymb}
\usepackage{algorithm2e}
\usepackage{multirow}
\usepackage{tikz, subfigure, multirow}
\usepackage{color}
\usepackage{xcolor}
\usepackage{mathtools, nccmath}
\usepackage{hhline}
\usepackage{makecell}
\usepackage{bm}
\usepackage[percent]{overpic}
\usepackage[outline]{contour}
\usetikzlibrary{positioning, arrows, fit, backgrounds, calc, shapes.misc, shapes, automata}

\newcommand\blfootnote[1]{%
  \begingroup
  \renewcommand\thefootnote{}\footnote{#1}%
  \addtocounter{footnote}{-1}%
  \endgroup
}

\usepackage[pagebackref=true,breaklinks=true,letterpaper=true,colorlinks,bookmarks=false]{hyperref}

\cvprfinalcopy 

\def\cvprPaperID{572} 

\ifcvprfinal\fi
\begin{document}

\title{CascadePSP: Toward Class-Agnostic and Very High-Resolution Segmentation via Global and Local Refinement}

\author{
\begin{tabular}{cc}
Ho Kei Cheng$^*$ & Jihoon Chung$^*$
\end{tabular}\\
HKUST\\
{\tt\small \{hkchengad,jchungaa\}@cs.ust.hk}
\and
Yu-Wing Tai\\
Tencent\\
{\tt\small yuwingtai@tencent.com}
\and
Chi-Keung Tang\\
HKUST\\
{\tt\small cktang@cs.ust.hk}
}

\maketitle

\begin{abstract}
State-of-the-art semantic segmentation methods were almost exclusively trained on images within a fixed resolution range. These segmentations are inaccurate for very high-resolution images since using bicubic upsampling of low-resolution segmentation does not adequately capture high-resolution details along object boundaries.
In this paper, we propose a novel approach to address the high-resolution segmentation problem \underline{without} using any high-resolution training data. 
The key insight is our CascadePSP network which refines and corrects local boundaries whenever possible. 
Although our network is trained with low-resolution segmentation data, our method is applicable to any resolution even for very high-resolution images larger than 4K.
We present quantitative and qualitative studies on different datasets to show that CascadePSP can reveal pixel-accurate segmentation boundaries using our novel refinement module \underline{without} any finetuning. Thus, our method can be regarded as class-agnostic.
Finally, we demonstrate the application of our model to scene parsing in multi-class segmentation.
\end{abstract}

\vspace{-2em}

\blfootnote{
$^*$Equal contribution.
This research is supported in part by Tencent and the Research Grant Council of the Hong Kong SAR under grant no. 1620818.}
\section{Introduction}
Resolution of commodity cameras and displays has significantly increased with 4K UHD ($3840\times2160$) being the high industry standard. Despite the demand for high-resolution media, many state-of-the-art computer vision algorithms face various challenges with images with high pixel count. 
Image semantic segmentation is one of these computer vision tasks. 
Models for semantic segmentation in deep learning designed for low-resolution images (\eg PASCAL or COCO dataset) often fail to generalize to higher resolution scenarios. 
Specifically, these models typically use GPU memory linear to the number of pixels, making it practically impossible to directly train a 4K UHD segmentation. 
High-resolution training data for semantic segmentation is difficult to obtain because pixel-accurate annotation is required, much less that even such high-resolution training data are available, to train a model on very high-resolution images, a much larger receptive field is required to capture sufficient semantics. Plausible workarounds include downsampling and cropping, but the former removes details while the latter destroys image context.

\begin{figure}[t]
	\includegraphics[width=\linewidth]{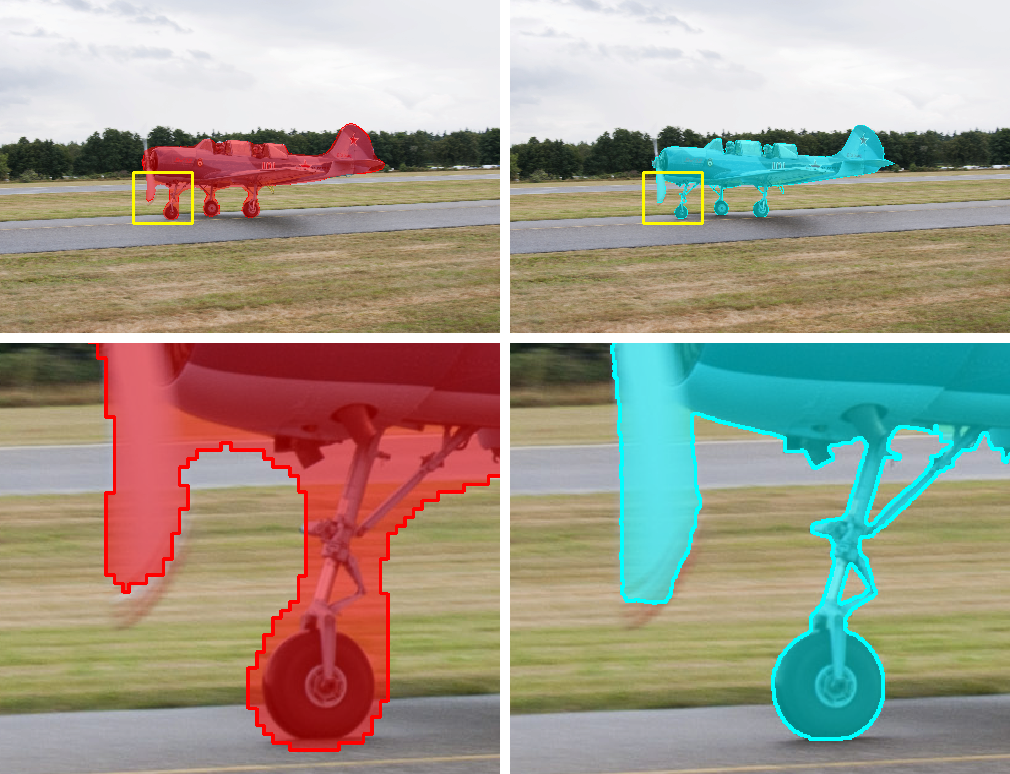}
	\label{fig:teaser}
	\caption{
	Segmentation of a high-resolution image ($3492 \times 2328$). \textbf{Left}: Produced by Deeplab V3+~\cite{chen2018encoder}. \textbf{Right}: Refined by our algorithm.}
	\vspace{-0.15in}
\end{figure}

This paper proposes \textbf{CascadePSP}\footnote{Source code,  pretrained models and dataset are available at \url{https://github.com/hkchengrex/CascadePSP}.}, a general segmentation refinement model that refines any given segmentation from low to high resolution. 
Our model is trained independently and can be easily appended to any existing methods to improve their segmentation, a finer and more accurate segmentation mask of an object can be produced. 
Our model takes as input an initial mask that can be an output of any algorithm to provide a rough object location. Then our CascadePSP will output a refined mask.
Our model is designed in a cascade fashion that generates refined segmentation in a coarse-to-fine manner. Coarse outputs from the early levels predict object structure which will be used as input to the latter levels to refine boundary details. 
Figure~\ref{fig:teaser} shows that the model not only generates output segmentation in very high-resolution but also refines and corrects erroneous boundary to produce more accurate result. 

To evaluate on very high-resolution images, we have annotated a high-resolution dataset with 50 validation and 100 test objects with the same semantic classes as in PASCAL, dubbed the “BIG” dataset. 
We test our model on PASCAL VOC 2012, BIG, and ADE20K. With a single model {\em without} using the dataset itself for finetuning, we have achieved consistent improvement over the state-of-the-art methods across these datasets and models. 
We show that our model does not have to be trained with respect to a specific dataset, or with outputs of a specific model. Rather, performing data augmentation by perturbing the ground truth is sufficient. 
We also show that our model can be extended to scene parsing for dense multi-class semantic segmentation with straightforward adaptation.
Our main contributions can be summarized as:

\vspace{-5pt}
\begin{itemize}
    \item We propose CascadePSP, a general cascade segmentation refinement model that can refine any given input segmentations, boosting the performance of state-of-the-art segmentation models without finetuning.
    \vspace{-5pt}
    \item We further show that our method can be used to produce high-quality and very high-resolution segmentations which has never been achieved by previous deep learning based methods.
    \vspace{-5pt}
    \item We introduce the BIG dataset that can be used as an accurate evaluation dataset for very high resolution semantic image segmentation task. 
\end{itemize}

\vspace{-0.10in}
\section{Related Works}
\vspace{-0.05in}
\paragraph{Semantic Segmentation}
Fully Convolutional Neural Networks (FCN) was first introduced in semantic segmentation in~\cite{long2015fully} which achieved remarkable progress at the time of introduction.
While FCNs capture information from bottom-up, contextual information with wide field-of-view is also important for pixel labeling tasks and is exploited by many segmentation models~\cite{chen2017deeplab, chen2016attention, farabet2012learning, he2004multiscale, Mostajabi2015feedforward, shotton2009textonboost, zhang2018context}, including image pyramid methods that use multi-scale inputs~\cite{chen2016attention, Dai2015ConvolutionalFM,farabet2012learning, lin2017refinenet, lin2016efficient, pinheiro2014recurrent}, or feature pyramid methods that use feature maps of different receptive field sizes by spatial pooling~\cite{liu2015parsenet, zhao2017pyramid} or dilated convolutions with different rates~\cite{chen2017deeplab, chen2017rethinking, chen2018encoder, li2018pyramid, wang2018understanding, yu2015multi}.
We choose PSPNet~\cite{zhao2017pyramid} for pyramid pooling in our network because the pertinent module is independent of input resolution, thus providing a simple yet effective method to capture contextual information even when the training and testing resolution significantly differ as in our case.

Encoder-decoder models have also been widely used in segmentation methods~\cite{badrinarayanan2015segnet, chen2018encoder, li2018pyramid, liu2019auto, liu2018path, noh2015learning, ronneberger2015u, wang2018understanding}. They first reduce the spatial dimension to capture high-level semantics and then recover the spatial extent using a decoder. Skip connections~\cite{drozdzal2016importance, ronneberger2015u, sun2019high} can be added to produce sharper boundaries which we have also employed.

Semantic segmentation models typically have a large output stride such as 4 or 8~\cite{chen2014semantic, chen2017deeplab, chen2017rethinking} due to memory and computational limitation. Outputs with stride are usually bilinearly upsampled to the target size, leading to inaccurate boundary labels. Recently, the authors of~\cite{chen2019collaborative} have proposed Global-Local Networks (GLNet) to solve this problem using a global information branch with a local fine structure network. However, they still require high-resolution training images which are not available for most tasks.

Our method adopts the encoder-decoder model to obtain better semantic and boundary information with a refinement cascade, which also helps to efficiently generate high-resolution segmentations. This formulation also makes our method highly robust and can generalize to high-resolution data without finetuning.

\noindent \textbf{Refining Segmentation} \space\space
FCN based methods typically do not generate very high-quality segmentation. Researchers have addressed this issue with graphical models such as CRF~\cite{chen2014semantic, chen2017deeplab, lin2016efficient, krahenbuhl2011efficient, liu2015semantic, zheng2015conditional} or region growing~\cite{dias2018semantic}. They often adhere to low-level color boundaries without fully leveraging high-level semantic information and cannot fix large error regions.
Propagation-based approaches~\cite{liu2017learning} cannot handle very high-resolution data due to computational and memory constraints.
Separate refinement modules are also used to increase boundary accuracy~\cite{peng2017large, xu2017deepmatting, zhang2019canet}.
They are trained in an end-to-end fashion. Large models are prone to overfitting~\cite{zhang2019canet} while shallow refinement networks~\cite{peng2017large,xu2017deep} have limited refinement capability. 
Contrary, our method has a high model capacity and can be trained independently to repair segmentation using only objectness. Finetuning with the specific model is not required so our training is not hindered by overfitting. 

 
\noindent \textbf{Cascade Network} \space \space
Multi-scale analysis leverages both large and small scale features in many computer vision tasks, such as edge detection~\cite{he2019bi, xie2015holistically}, detection~\cite{lin2017feature, liu2016ssd, sun2013deep}, and segmentation~\cite{chen2019collaborative, lin2017refinenet, zhao2018icnet}. 
In particular, a number of methods~\cite{lin2017refinenet, xie2015holistically, zhao2018icnet} predict independent results at each stage and merge them to obtain multi-scale information. 
Our method not only fuses features from coarse scales but uses them as one of the inputs for the next finer level. We will show that adding coarse outputs as input for the next level does not change our formulation and thus the same network can be used recursively for higher resolution refinement.

\vspace{-0.10in}
\section{CascadePSP}
\vspace{-0.05in}
In this section, we first describe our single refinement module and then our cascade method which makes use of multiple refinement modules for high-resolution segmentation.

\vspace{-0.05in}
\subsection{Refinement Module}
\begin{figure}[t]
	\begin{center}
		\includegraphics[width=\linewidth]{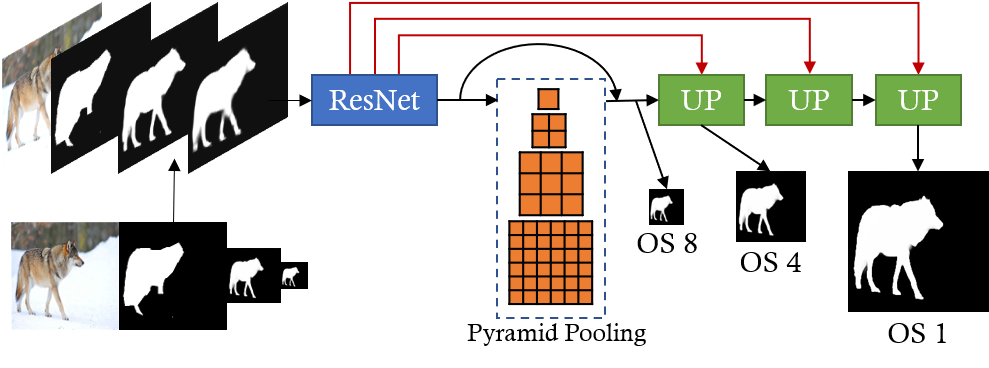}
	\end{center}
	\vspace{-0.15in}
	\caption{\textbf{Refinement module} (RM). Network structure of a single RM, taking three levels of segmentation as inputs to refine the segmentation with different output strides (OS) in different branches. Red lines denote skip-connections. In this paper, we use output strides of 8, 4, and 1.}
	\label{fig:single_level}
	\vspace{-0.15in}
\end{figure}

\vspace{-0.05in}
As illustrated in Figure~\ref{fig:single_level}, our refinement module takes an image and multiple imperfect segmentation masks at different scales to produce a refined segmentation. 
Multi-scale inputs allow the model to capture different levels of structural and boundary information, which allow the network to learn to adaptively fuse the mask features from different scales to refine the segmentation at the finest level. 

All the input segmentations at lower resolution are bilinearly upsampled to the same size and concatenated with the RGB image. We extract stride~8 feature maps from the inputs using PSPNet~\cite{zhao2017pyramid} with ResNet-50~\cite{he2016deep} as the backbone. We follow the pyramid pooling sizes of $[1, 2, 3, 6]$ as in~\cite{zhao2017pyramid} which helps to capture global context. Besides the final stride~1 output, our model also generates intermediate stride~8 and stride~4 segmentations which focus on fixing the overall structure of the input segmentation. We skip stride 2 to provide flexibility to correct local error boundary.
 
To reconstruct pixel-level image details that are lost in the extraction process, we employ skip-connection from the backbone network and fuse the features using an upsampling block. We concatenate the skip connected features and the bilinearly upsampled features from the main branch, and process them with two ResNet blocks. A segmentation output is generated using a 2-layer $1\times1$ conv followed by a sigmoid activation. 

\noindent \textbf{Loss} \space \space
We produce the best result using cross-entropy loss for the coarser stride~8 output, L1+L2 loss for the finer stride~1 output, and the average of cross-entropy and L1+L2 loss for the intermediate stride~4 output. 
Different loss functions are applied for different strides because the coarse refinement focuses on the global structure while ignoring local details, while the finest refinement aims to achieve pixel-wise accuracy by relying on local cues.
To encourage better boundary refinement, L1 loss on segmentation gradient magnitude is also employed on the stride~1 output. The segmentation gradient is estimated by a $3\times3$ mean filter followed by a Sobel operator~\cite{kanopoulos1988design}. The gradient loss makes outputs adhere better to the object boundary at the pixel level. As gradient is sparser compared to pixel level loss, we weigh it with $\alpha$, which is set to 5 in our experiments. The gradient loss can be written as:
\begin{equation*}
\mathcal{L}_{\mathit{grad}} = \alpha \cdot \frac{1}{n} \sum_i{\left\lVert\nabla(f_m(x_i))-\nabla(f_m(y_i))\right\rVert_1}
\end{equation*}
where $f_m(\cdot)$ denotes the $3\times3$ mean filter, $\nabla$ denotes the gradient operator approximated by a Sobel operator, $n$ is the total number of pixels, $x_i$ and $y_i$ are the $i$th pixel of the ground truth segmentation and output segmentation respectively. Our final loss can be written as:
\begin{equation*}
\mathcal{L} = \mathcal{L}^8_{CE} + \frac{1}{2} (\mathcal{L}^4_{L1+L2} + \mathcal{L}^4_{CE}) + \mathcal{L}^1_{L1+L2} + \mathcal{L}^1_{\mathit{grad}}
\end{equation*}
where $\mathcal{L}^s_{CE}$, $\mathcal{L}^s_{L1+L2}$, and $\mathcal{L}^s_{\mathit{ grad}}$ denote cross-entropy loss, L1+L2 loss, and gradient loss for output stride $s$ respectively. 

\noindent \textbf{Ablation Study of Refinement Module} \space \space
We evaluate our method using standard segmentation metric IoU. To highlight the perceptual importance of boundary accuracy, we propose a new \textbf{m}ean \textbf{B}oundary \textbf{A}ccuracy measure (\textbf{mBA}). 
For a robust estimation for images of different sizes, we sample $5$ radii in $[3, \frac{w+h}{300}]$ with uniform intervals, compute the segmentation accuracy within each radius from the ground truth boundary, then average these values.
Here we perform ablation studies to show the efficacy of our cascade design and loss function. Table~\ref{tab:ablation_rm} shows that our model produces the most significant improvement in IoU and even more significantly in boundary accuracy.

\begin{table}[t]
\centering
\begin{tabular}{l|c|c}
  \multirow{2}{*}{Configuration} & \multicolumn{2}{c}{\textbf{PASCAL VOC 2012}}\\
  & IoU (\%) & mBA (\%)\\
  \hhline{|=|=|=|}
  Deeplab V3+ & 87.13 & 61.68 \\
  \hline
  \multicolumn{3}{c}{\textbf{Ablation of network structure}} \\
  \Xhline{2\arrayrulewidth}
  Vanilla FCN & $88.46_{\uparrow1.33}$ & $70.38_{\uparrow8.70}$ \\
  With OS1 only & $88.76_{\uparrow1.63}$ & $71.49_{\uparrow9.81}$ \\
  With OS8 \& OS1 only & $88.85_{\uparrow1.72}$ & $71.88_{\uparrow10.2}$ \\
  \hline
  \multicolumn{3}{c}{\textbf{Ablation of loss function}} \\
  \Xhline{2\arrayrulewidth}
  CE loss only & $88.73_{\uparrow1.60}$ & $71.07_{\uparrow9.39}$ \\
  L1+L2 loss only & $88.74_{\uparrow1.61}$ & $71.07_{\uparrow9.39}$ \\
  CE and L1+L2 loss only  & $88.84_{\uparrow1.71}$ & $71.36_{\uparrow9.68}$ \\
  \Xhline{3\arrayrulewidth}
  Ours - Final & $\mathbf{89.01_{\uparrow1.88}}$ & $\mathbf{72.10_{\uparrow10.4}}$ \\
\end{tabular}
\caption{Ablation study of the refinement module. With the proposed 3-level cascade and loss function, we achieve the highest gain over the input segmentation model.}
\label{tab:ablation_rm}
\vspace{-0.15in}
\end{table}

\begin{figure}[t]
	\begin{center}
		\includegraphics[width=0.8\linewidth]{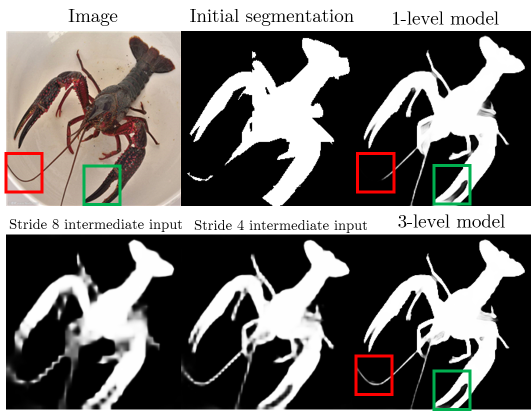}
	\end{center}
	\vspace{-0.15in}
	\caption{Difference between a 3-level input model and a 1-level input model. The 3-level input model uses small-scale intermediates (bottom row, left two) that, though inaccurate, capture structural information (e.g. the tentacle) to be refined at the later stage.}
	\label{fig:1vs3}
	\vspace{-0.15in}
\end{figure}

With a multi-level cascade, the module can delegate different stages of refinement to different scales. 
As shown in Figure~\ref{fig:1vs3},  the 3-level model uses intermediate small-scale segmentations (will be detailed in Section~\ref{sec:cascade}) to better capture object structure. 
Although both models have the same receptive field, the 3-level model can better leverage structural cues to produce a more detailed segmentation than the 1-level model.

\subsection{Global and Local Cascade Refinement} \label{sec:cascade}
In testing, we use the \textbf{Global step} and the \textbf{Local step} to perform high-resolution segmentation refinement by employing the {\em same} trained refinement module. Specifically, the Global step considers the whole resized image to repair structure while the Local step refines details in full resolution using image crops. The same refinement module can be used recursively for higher resolution refinement.

\begin{figure}[t]
\begin{center}
    \includegraphics[width=\linewidth]{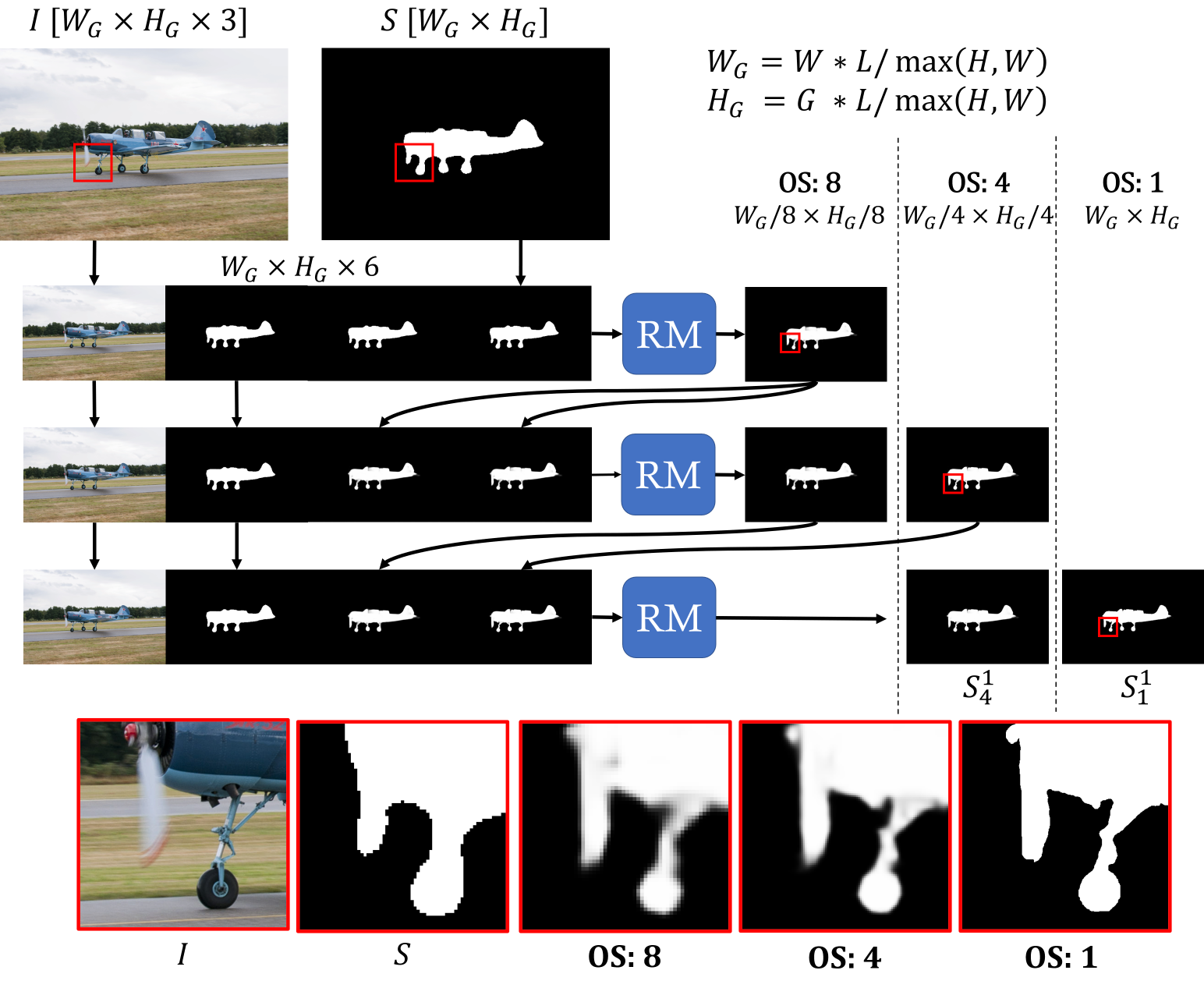}
\end{center}
    \vspace{-0.15in}
	\caption{\textbf{Global step} refines the whole image using the same refinement module (RM) to perform a 3-level cascade with output strides (OS) of 8, 4, and 1. The cascade is jointly optimized, capturing object structure at large output strides and accurate boundary at small output strides (i.e., with a higher resolution).
	}
	\label{fig:global_fig}
	\vspace{-0.15in}
\end{figure}

\subsubsection{Global Step}
Figure~\ref{fig:global_fig} details the design of the Global step which refines the whole image with a 3-level cascade. As the full-resolution image during testing often cannot be fit into the GPU for processing, we downsample the input such that the long-axis has length $L$ while maintaining the same aspect ratio. 

Inputs to the cascade are initialized with the input segmentation, which is replicated to keep the input channel dimension
constant. 
After the first level of the cascade, one of the input channels will be replaced with the bilinearly upsampled coarse output. 
This is repeated until the last level, where the input consists of both the initial segmentation and all outputs from previous levels. 

This design enables our network to fix segmentation errors progressively while keeping details present in the initial segmentation. 
With multiple levels, we can roughly delineate the object and fix larger error in coarse levels, and focus on boundary accuracy in fine levels using more robust features provided by the coarse levels. 

\begin{figure}[t]
\begin{center}
    \includegraphics[width=\linewidth]{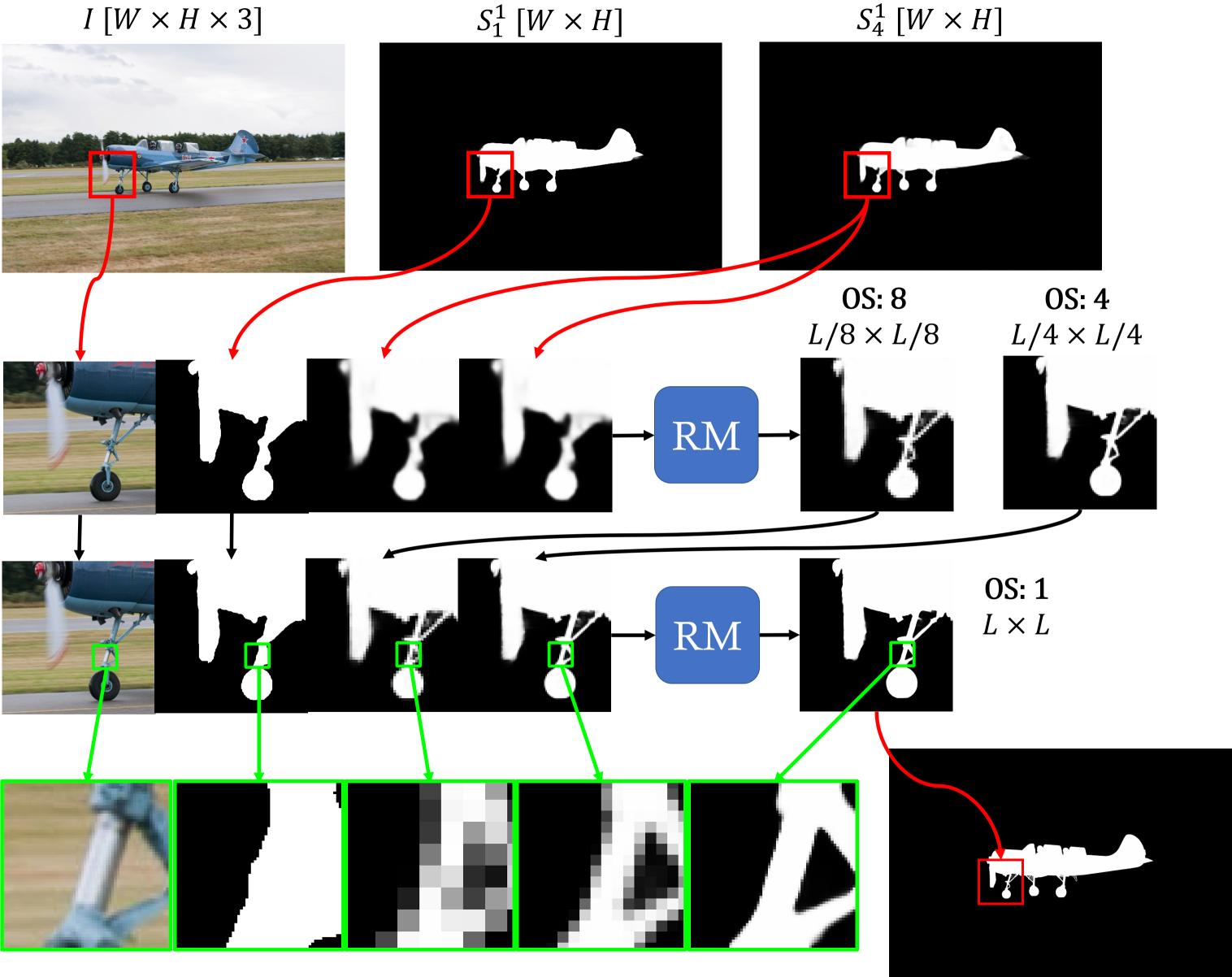}
\end{center}
    \vspace{-0.15in}
	\caption{\textbf{Local step} takes the outputs from the Global step, and feeds them through a 2-level cascade constructed with the same refinement module with output strides of 4 and 1 respectively. This figure shows the process for a single image crop as shown by the red lines, and green lines show visual improvements of our refinement. Outputs from all the image crops will be fused as the final output. 
	}
	\label{fig:local_fig}
	\vspace{-0.3in}
\end{figure}

\subsubsection{Local Step}
Figure~\ref{fig:local_fig} illustrates the details of the Local step.
Very high-resolution images cannot be processed in a single pass even with modern GPUs due to the memory constraint. Also, the drastic change of scale between training and testing data will cause poor segmentation quality. We leverage our cascade model to first perform global refinement using a downsampled image, and then perform local refinement using image crops {\em from a higher resolution image}. These crops enable the Local step to handle high-resolution images without high-resolution training data while taking image context into account due to the Global step.

During the Local step, the model takes the two outputs of the last level of the Global step, denoted as $S^1_4$ and $S^1_1$. Both outputs are bilinearly resized to the original size of the image $W\times H$.
The model takes image crops of size $L\times L$ and $16$ pixels will be chipped away from each side of the crop output to avoid boundary artifacts, with exceptions at the image border. The crops are taken uniformly with a stride of $L/2-32$ such that most pixels are covered by four crops, and invalid crops that go beyond image borders are shifted to align with the last row/column of the image.
The image crops are then fed into a 2-level cascade with output stride of 4 and 1 respectively.
In fusion, the outputs from different patches might disagree with each other due to different image context, and we resolve this by averaging all the output values. 
For images with even higher resolution, we can apply the local step recursively in a coarse-to-fine manner. 

\begin{figure}[h]
\centering
	\includegraphics[width=\linewidth]{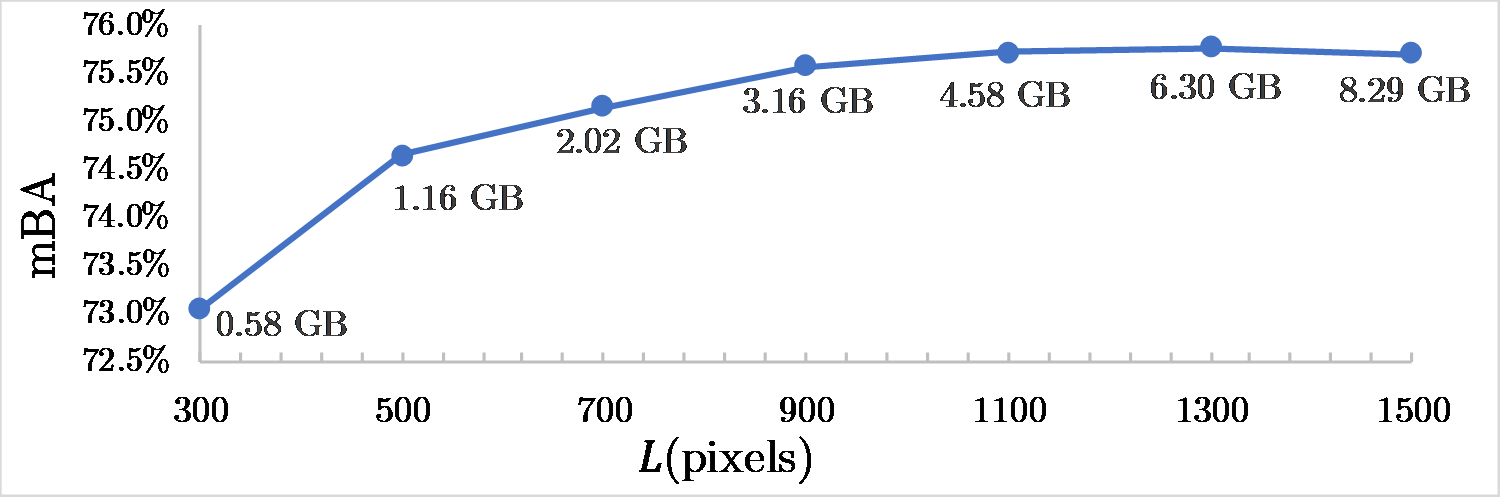}
	\caption{Relationship between the choice of $L$ and mBA in the BIG validation set. With increasing $L$, GPU memory usage increases with diminishing performance gain. }
	\label{fig:mem_iou}
	\vspace{-0.15in}
\end{figure}

\subsubsection{Choosing $L$} 
Figure~\ref{fig:mem_iou} presents the relationship between GPU memory usage and refinement quality (mBA) during testing when different $L$ is chosen. We have chosen $L=900$ with 3.16 GB of GPU memory usage in our experiments to balance the tradeoff between increasing GPU memory usage and diminishing performance gain.
Using an even higher $L$ is unnecessary and occupies extra memory in our experiments on the BIG validation set. 
In low-memory settings, a smaller $L$ such as $500$ can be used to produce a slightly worse ($-0.6$\% mBA) refinement with a much lower memory footprint (1.16 GB). Note that the GPU memory usage only relates to $L$ but not the image resolution as the fusion step can be easily performed on the CPU.

\subsubsection{Ablation Study for Global and Local Refinement}
Table~\ref{tab:ablation_global_local} shows that both the Global step and the Local step are essential to high-resolution segmentation refinement. 
Note that the IoU drop is much more significant when we remove the Global step, indicating that the Global step is mainly responsible for fixing the overall structure contributing more to IoU boost while the Local step alone cannot achieve due to insufficient image context. Without the Local step, although IoU only decreases slightly, we note that boundary accuracy decreases more significantly since the Global step cannot extract high-resolution details.

\begin{figure}[t]
\begin{center}
    \includegraphics[width=\linewidth]{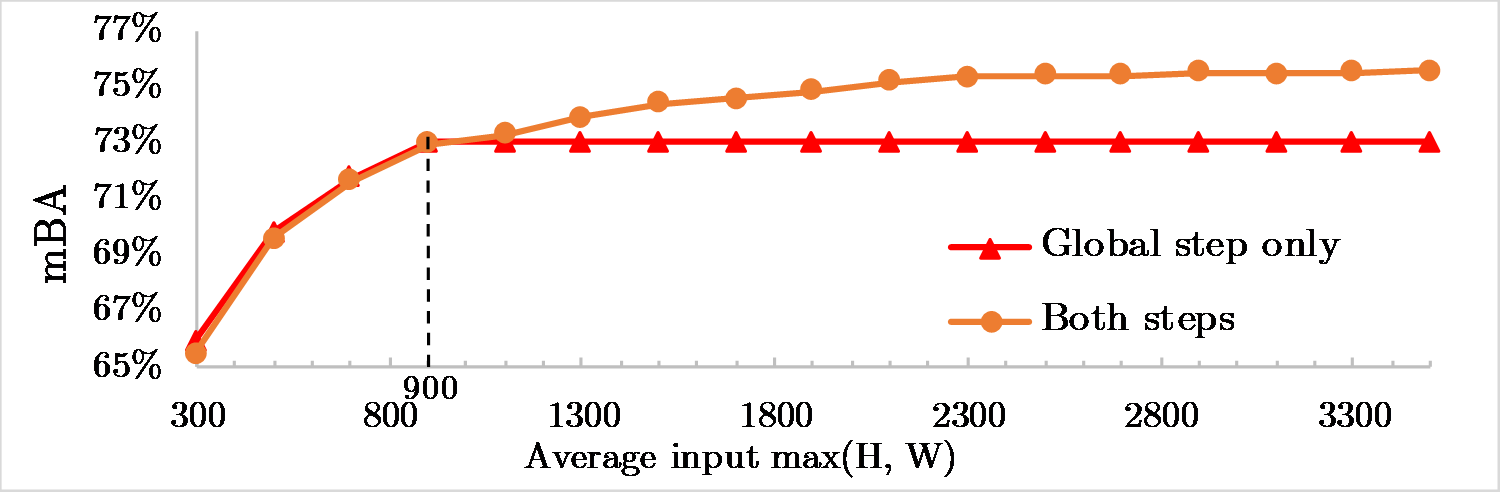}
\end{center}
    \vspace{-0.15in}
    \caption{
	We evaluated our method across different input resolutions. The Global step does not benefit from higher resolution inputs because it is bounded by $L=900$. The Local step is crucial for high-resolution refinement to handle inputs with a larger size than $L$ with bounded memory cost.
	}
	\label{fig:global_only_and_both_steps}
	\vspace{-0.15in}
\end{figure}

Figure~\ref{fig:global_only_and_both_steps} studies the importance of the Local step for different resolution inputs: we evaluate our method with and without the Local step in various-sized segmentations generated by resizing the BIG validation set.
While the Global step is sufficient for low-resolution inputs, the Local step is crucial for accurate high-resolution refinement with size higher than the switching point $900$.
We therefore use both the Global and Local step for inputs with $\max(H, W)\geq 900$, and only the Global step for lower resolution inputs.

\vspace{-0.20in}
\begin{table}[ht]
\centering
\begin{tabular}{l|c|c}
  \multirow{2}{*}{Configuration} & \multicolumn{2}{c}{\textbf{BIG}}\\
  & IoU (\%) & mBA (\%)\\
  \hhline{|=|=|=|}
  Deeplab V3+ & 89.65 & 60.94 \\
  \hline
  Global step only & $91.86_{\uparrow2.21}$ & $73.10_{\uparrow12.2}$ \\
  Local step only & $91.35_{\uparrow1.70}$ & $73.06_{\uparrow12.1}$ \\
  Both steps (Ours) & $\mathbf{92.01_{\uparrow2.36}}$ & $\mathbf{75.59_{\uparrow14.7}}$ \\
\end{tabular}
\caption{Ablation experiments for the Global and Local step. Input segmentations are taken from DeepLab V3+ on the BIG validation set. Using both steps show the best results.}
\label{tab:ablation_global_local}
\vspace{-0.15in}
\end{table}

\subsection{Training}
To learn objectness information, we train our model on a collection of datasets in a class-agnostic manner. We merge MSRA-10K~\cite{ChengPAMI}, DUT-OMRON~\cite{yang2013saliency}, ECSSD~\cite{shi2015hierarchical}, and FSS-1000~\cite{FSS1000} to generate a segmentation dataset of 36,572 with much more diverse semantic classes than common datasets such as PASCAL (20 classes) or COCO (80 classes).
Using this dataset ($>1000$ classes) makes our model more robust and generalizable to new classes.

During training, we take random $224\times224$ image crops and generate input segmentations by perturbing the ground truth. The inputs go through a 3-level cascade as in the Global step with the loss computed in every level. 
Although the crop size is smaller than $L$ which is used in testing, our model design helps bridging this gap. The fully convolutional feature extractor provides translational invariance while the pyramid pooling module provides important image context, allowing our model to be extended to higher resolution without significant performance loss. The use of smaller crop speeds up our training process and makes data preparation much easier as high-resolution training data for segmentation is expensive to obtain.

\begin{figure}[t]
	\minipage[t]{0.24\linewidth}
	\includegraphics[width=\linewidth]{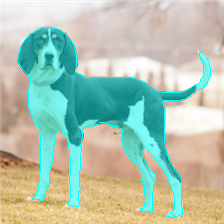}
	\endminipage\hfill
	\minipage[t]{0.24\linewidth}
	\includegraphics[width=\linewidth]{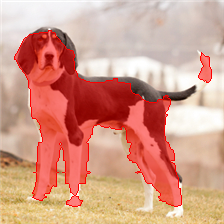}
	\endminipage\hfill
	\minipage[t]{0.24\linewidth}
	\includegraphics[width=\linewidth]{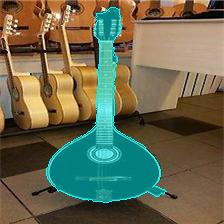}
	\endminipage\hfill
	\minipage[t]{0.24\linewidth}
	\includegraphics[width=\linewidth]{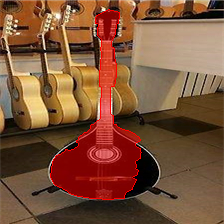}
	\endminipage\hfill
	\caption{\textbf{Blue}: Ground truth labels of FSS-1000~\cite{FSS1000}.
	\textbf{Red}: Perturbed labels that we use as inputs to train our model.}
	\label{fig:perturb}
	\vspace{-0.15in}
\end{figure}

For generalizability, we avoid training using segmentation outputs generated by existing models which can lead to overfitting to that specific model. 
Instead, perturbed ground truth should portray various shapes and output of inaccurate segmentations produced by other methods, which helps our algorithm to be more robust to different initial segmentations. We generate such perturbed segmentations by subsampling the contour followed by random dilations and erosions. Examples of such perturbation are shown in Figure~\ref{fig:perturb}.

\section{Experiments}
In this section, we quantitatively evaluate our results with PASCAL VOC 2012 \cite{Pascal}, BIG (our high-resolution data set), and ADE20K \cite{zhou2017scene}. We evaluate our model \textit{without any finetuning} in various settings and show the improvements made by our model.

\subsection{Dataset and Evaluation Method} \label{sec:eval_method}
\begin{figure}[t]
\centering
	\minipage[t]{0.24\linewidth}
	\includegraphics[width=\linewidth]{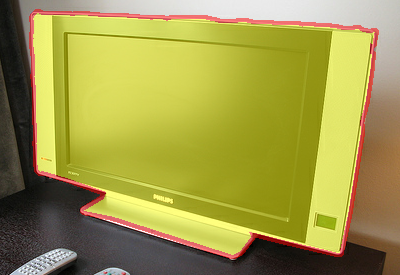}
	\endminipage \hfill
	\minipage[t]{0.24\linewidth}
	\includegraphics[width=\linewidth]{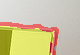}
	\endminipage \hfill
	\minipage[t]{0.24\linewidth}
	\includegraphics[width=\linewidth]{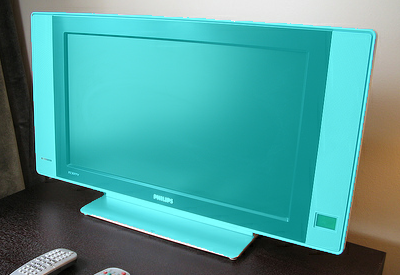}
	\endminipage \hfill
	\minipage[t]{0.24\linewidth}
	\includegraphics[width=\linewidth]{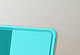}
	\endminipage \hfill
	\caption{Segmentation results in the PASCAL VOC 2012 validation set. \textbf{Left}: An example ground truth label of PASCAL VOC 2012. Red line shows the void boundary label.
	\textbf{Right}: Relabeled segmentation for the same image.}
	\label{fig:fine}
	\vspace{-0.15in}
\end{figure}

Although widely used in image segmentation tasks, PASCAL VOC 2012 dataset does not have pixel-perfect segmentations and the areas near the boundary are labeled as ``void''. For a more accurate evaluation, we have relabeled 500 segmentations from the PASCAL VOC 2012 validation set, so that the accurate boundary can be found within the void boundary regions. Figure~\ref{fig:fine} shows a relabeled example.

The lack of a high-resolution image segmentation dataset is one of the difficulties of evaluating an image segmentation model in high-resolution. To solve this issue, we present the BIG dataset, a high-resolution semantic segmentation dataset with 50 validation and 100 test objects. Image resolution in BIG ranges from $2048 \times 1600$ to $5000 \times 3600$. Every image in the dataset has been carefully labeled by a professional while keeping the same guidelines as PASCAL VOC 2012 without the void region. 
Both the relabeled PASCAL validation set and the BIG dataset are available on our project website.
Other datasets used in the evaluation are not modified.
We evaluate our method using standard segmentation metric IoU and our boundary metric mBA.

\subsection{Implementation Details}
We implement our model with PyTorch~\cite{paszke2017automatic}. We use PSPNet with  ResNet-50 backbone~\cite{zhao2017pyramid} as our base network. Data augmentations, including perturbation of ground truths, image flipping and cropping are done on-the-fly to further increase data variety. We use Adam optimizer~\cite{kingma2014adam} with a weight decay of $10^{-4}$, learning rate of $3\times10^{-4}$ for 30K iterations followed by a learning rate of $3\times10^{-5}$ for another 30K iterations with a batch size of 9. The total training time is around 16h with two 1080Ti. The Local step is only performed in the region of interest, and the complete refinement process takes about 6.6s for Figure~\ref{fig:teaser}. 
Unless otherwise specified, we use the same trained model for all the experiments. 

\subsection{Segmentation Input}
Our method can refine input segmentations using only objectness information. Note that our model has \textit{never} seen any of the following datasets in training. In this section, we focus on the refinement effect of individual objects, meaning that class competition is not introduced. 

Here, we compare and evaluate the effect of our refinement model on the output of various semantic segmentation models~\cite{chen2018encoder, zhao2017pyramid} trained on the PASCAL VOC 2012 dataset. Our method is more effective than commonly used multi-scale testing, with experimental results further shown in the supplementary material.

\vspace{-0.1in}
\paragraph{PASCAL VOC 2012}
As the input models are trained in the PASCAL VOC 2012 dataset, 
resizing is not needed to obtain their outputs
which are then fed into our refinement model. These images are of low resolution, so we can refine them directly using the Global step only. 
We report the overall class-agnostic IoU and boundary accuracy in the upper half of Table~\ref{tab:semantic_experiment}. 
Results show that our method can improve segmentation quality in all cases, especially along the boundary region.

\begin{figure}[h]
\centering
	\includegraphics[width=\linewidth]{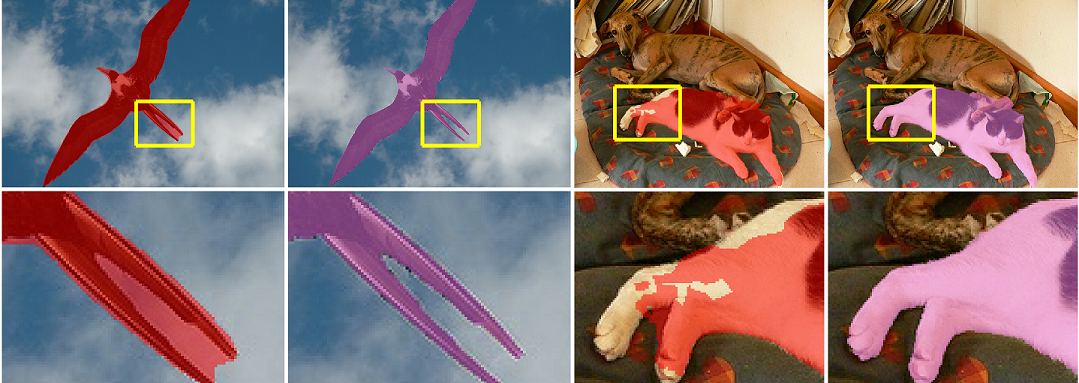}
	\caption{\textbf{Red}: Output produced by Deeplab V3+. 
	\textbf{Purple}: Segmentation refined by our algorithm.}
	\label{fig:pascal_fine_results}
	\vspace{-0.15in}
\end{figure}

\begin{table}[h]
	\centering
	\begin{tabular}{l|c|c}
		Methods & IoU (\%) & mBA (\%)\\
		\hhline{|=|=|=|} 
		\multicolumn{3}{c}{\textbf{PASCAL VOC 2012}} \\
		\Xhline{2\arrayrulewidth}
		FCN-8s~\cite{long2015fully} & 68.85  & 54.05 \\
		\quad (+) Ours & $\mathbf{72.70_{\uparrow3.85}}$ & $\mathbf{65.36_{\uparrow11.3}}$ \\
		\hline
		RefineNet~\cite{lin2017refinenet} & 86.21 & 62.61 \\
		\quad (+) Ours & $\mathbf{87.48_{\uparrow1.27}}$ & $\mathbf{71.34_{\uparrow8.73}}$ \\
		\hline
		DeepLabV3+~\cite{chen2018encoder} & 87.13 & 61.68 \\
		\quad (+) Ours & $\mathbf{89.01_{\uparrow1.88}}$ & $\mathbf{72.10_{\uparrow10.4}}$ \\
		\hline
		PSPNet~\cite{zhao2017pyramid} & 90.92 & 60.51 \\
		\quad (+) Ours & $\mathbf{92.86_{\uparrow1.94}}$ & $\mathbf{72.24_{\uparrow11.7}}$ \\
		\hhline{|=|=|=|} 
        \multicolumn{3}{c}{\textbf{BIG}} \\
		\Xhline{2\arrayrulewidth}
		FCN-8s~\cite{long2015fully} & 72.39  & 53.63 \\
		\quad (+) Ours & $\mathbf{77.87_{\uparrow5.48}}$ & $\mathbf{67.04_{\uparrow13.4}}$ \\
		\hline
		RefineNet~\cite{lin2017refinenet} & 90.20 & 62.03 \\
	    \quad (+) Ours & $\mathbf{92.79_{\uparrow2.59}}$ & $\mathbf{74.77_{\uparrow12.7}}$ \\
		\hline
		DeepLabV3+~\cite{chen2018encoder} & 89.42 & 60.25 \\
		\quad (+) Ours & $\mathbf{92.23_{\uparrow2.81}}$ & $\mathbf{74.59_{\uparrow14.3}}$ \\
		\hline
		PSPNet~\cite{zhao2017pyramid} & 90.49 & 59.63 \\
		\quad (+) Ours & $\mathbf{93.93_{\uparrow3.44}}$ & $\mathbf{75.32_{\uparrow15.7}}$ \\
	\end{tabular}
	\caption{Comparison between different semantic segmentation methods with and without our refinement. 
Their results are produced using their respective official implementations with the best provided model. Low-resolution outputs from the original model are bicubic-upsampled to the original resolution for evaluation.}
	\label{tab:semantic_experiment}
	\vspace{-0.15in}
\end{table}

\vspace{-0.1in}
\paragraph{BIG dataset}
Most existing segmentation methods cannot be directly evaluated on the full-resolution BIG dataset, due to the memory constraint. Therefore, we obtained initial segmentations by feeding resized images to the existing models. We downsampled the input image such that the long-axis is $512$-pixel while maintaining the aspect ratio, and bicubic-upsampled the output segmentation to the original resolution.

\begin{figure*}[t]
\begin{center}
    \begin{overpic}[width=\linewidth]{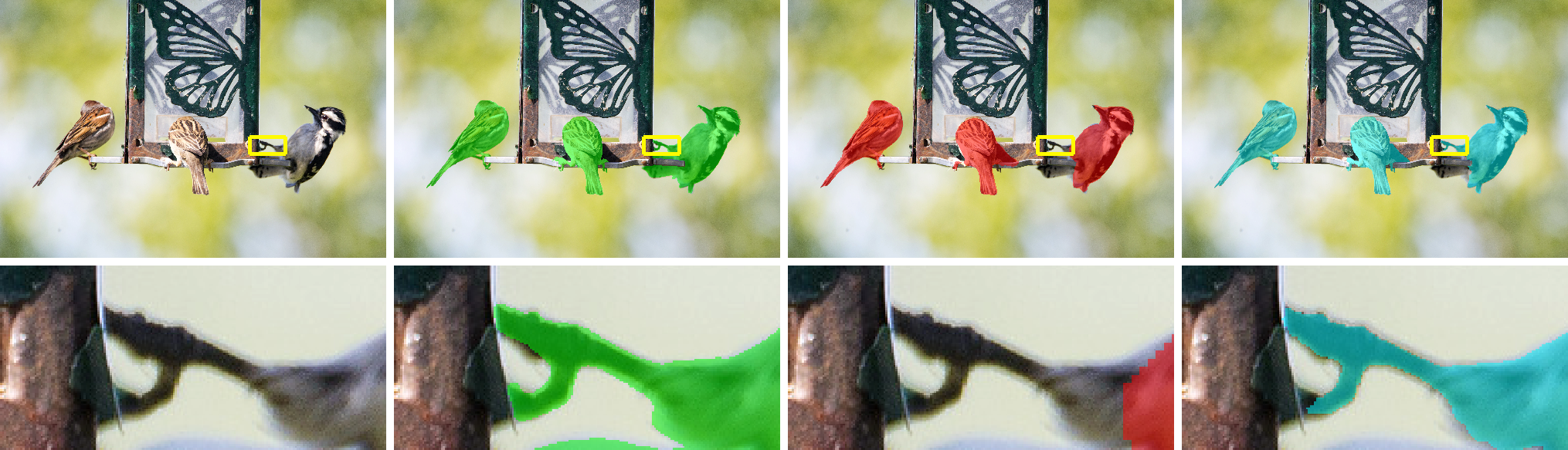}
     \put (51,26.5) {\contour{white}{DeeplabV3+}}
    \end{overpic}
    
	\vspace{0.05in}
    \begin{overpic}[width=\linewidth]{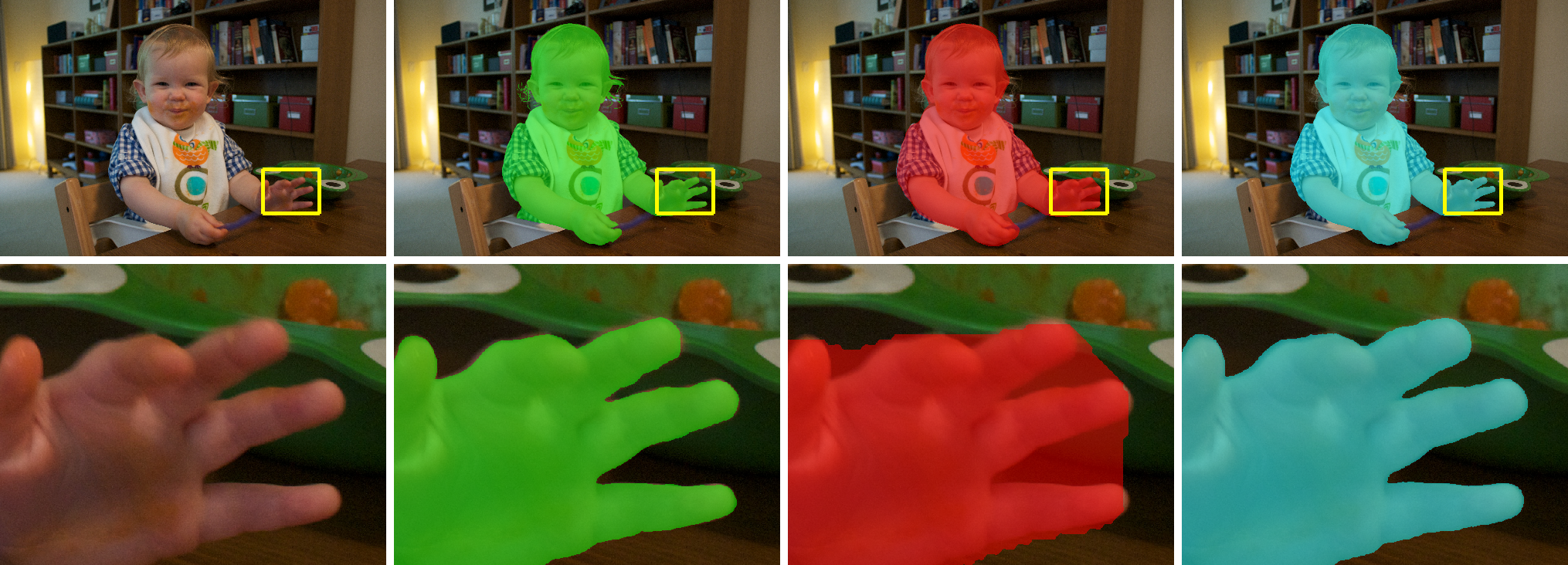}
     \put (51,34) {\contour{white}{RefineNet}}
    \end{overpic}
    
	\vspace{0.1in}
    \begin{overpic}[width=\linewidth]{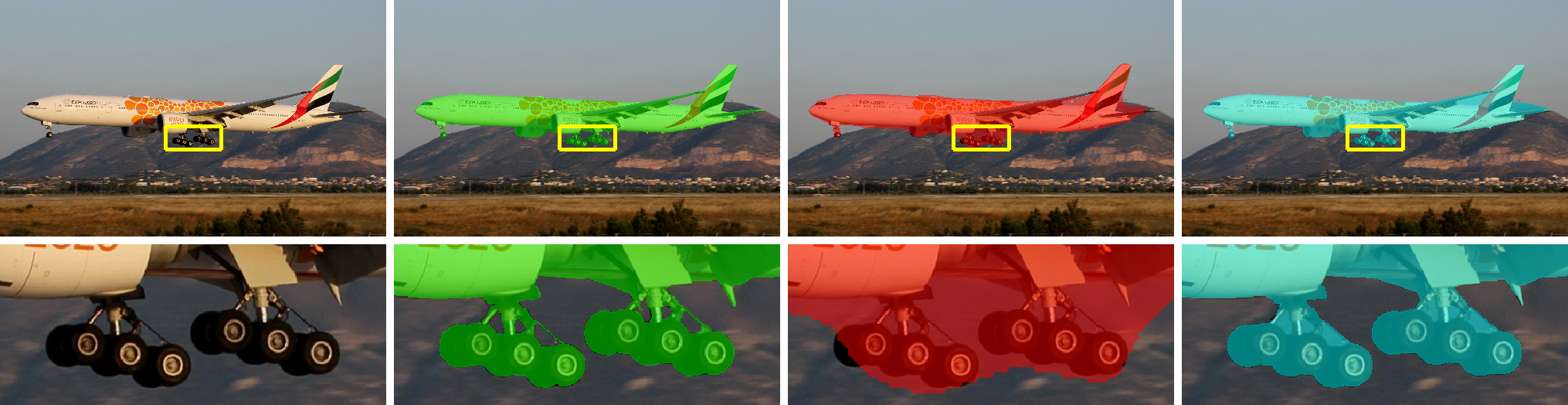}
     \put (51,23.5) {\contour{white}{PSPNet}}
    \end{overpic}
    
	\vspace{0.05in}
    \begin{overpic}[width=\linewidth]{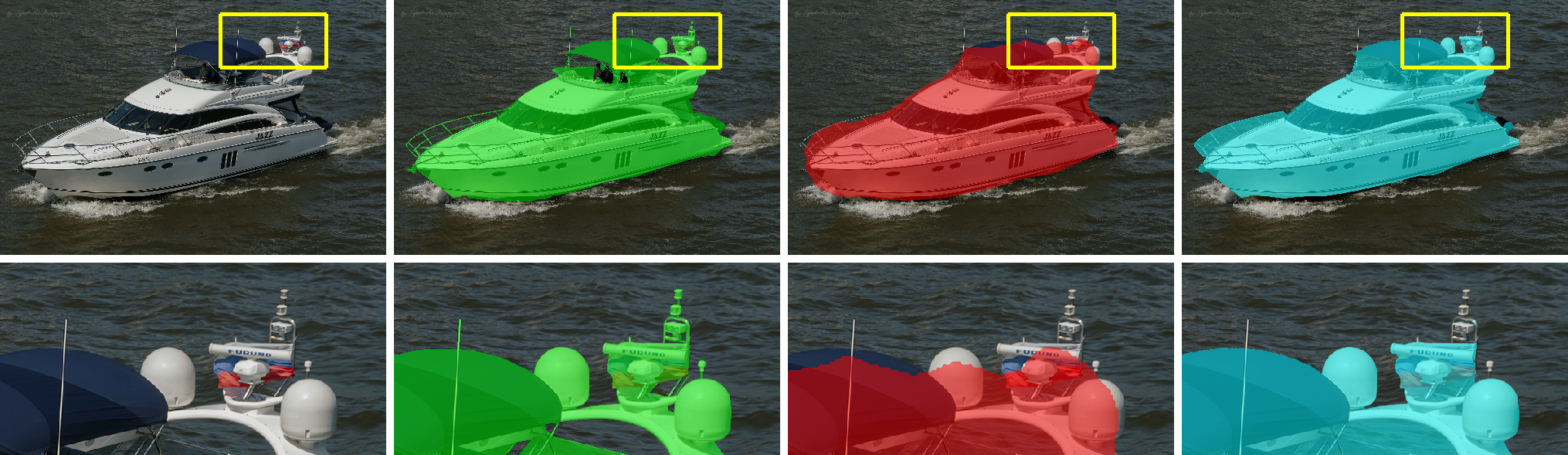}
     \put (51,27) {\contour{white}{FCN-8s}}
    \end{overpic}
	\minipage[t]{0.24\linewidth}
    	\begin{center}
        	\textbf{(a) Image}
    	\end{center}
	\endminipage\hfill
	\minipage[t]{0.24\linewidth}
    	\begin{center}
        	\textbf{(b) Ground Truth}
    	\end{center}
	\endminipage\hfill
	\minipage[t]{0.24\linewidth}
    	\begin{center}
        	\textbf{(c) Input}
    	\end{center}
	\endminipage\hfill
	\minipage[t]{0.24\linewidth}
    	\begin{center}
        	\textbf{(d) Our Refinement}
    	\end{center}
	\endminipage\hfill
\end{center}
   \caption{Visual comparison on the BIG test set. Odd rows show the whole image and even rows show the zoomed-in crop. Inputs are from DeeplabV3+, RefineNet, PSPNet, and FCN-8s, top to bottom.}
\label{fig:BIG_visual}
\end{figure*}

In the lower half of Table~\ref{tab:semantic_experiment}, we show our results on the BIG test set with high-resolution segmentation. Note that even we have never seen any high-resolution training images, we are able to produce high-quality refinements at these scales. Figure~\ref{fig:BIG_visual} shows the visual improvement of our refinement.
Although super-resolution models \cite{dong2015image, wang2018esrgan} may seem plausible for upsampling segmentation masks, inputs with erroneous segmentation (\eg the missing table leg in Figure~\ref{fig:ade_fine_results} and the baby's hand in Figure~\ref{fig:BIG_visual}) cannot be corrected by super-resolution.

\begin{figure}[t]
\centering
	\includegraphics[width=\linewidth]{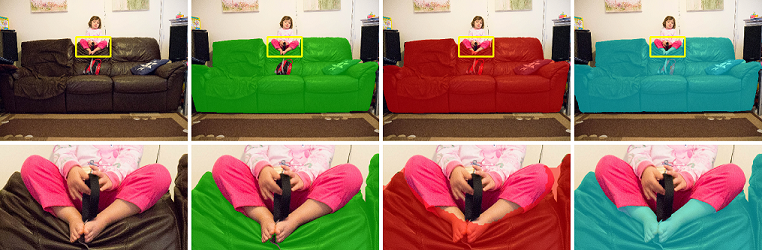}
	\minipage[t]{0.24\linewidth}
    	\begin{center}
        	\textbf{Image}
    	\end{center}
	\endminipage\hfill
	\minipage[t]{0.24\linewidth}
    	\begin{center}
        	\textbf{GT}
    	\end{center}
	\endminipage\hfill
	\minipage[t]{0.24\linewidth}
    	\begin{center}
        	\textbf{DeeplabV3+}
    	\end{center}
	\endminipage\hfill
	\minipage[t]{0.24\linewidth}
    	\begin{center}
        	\textbf{Ours}
    	\end{center}
	\endminipage\hfill
	\caption{A failure case of our method. DeeplabV3+ incorrectly labels a large region of the feet as foreground. Although our refinement still adheres well to the color boundary, it produces a wrong segmentation due to the lack of semantic information. }
	\label{fig:failure}
	\vspace{-0.15in}
\end{figure}

Our method relies on the input segmentation and low-level cues and does not have the specific semantic capability.
Figure~\ref{fig:failure} shows one failure case where the input error is too large for our method to eliminate.

\subsection{Scene parsing}
\begin{figure}
\centering
\includegraphics[width=0.75\linewidth]{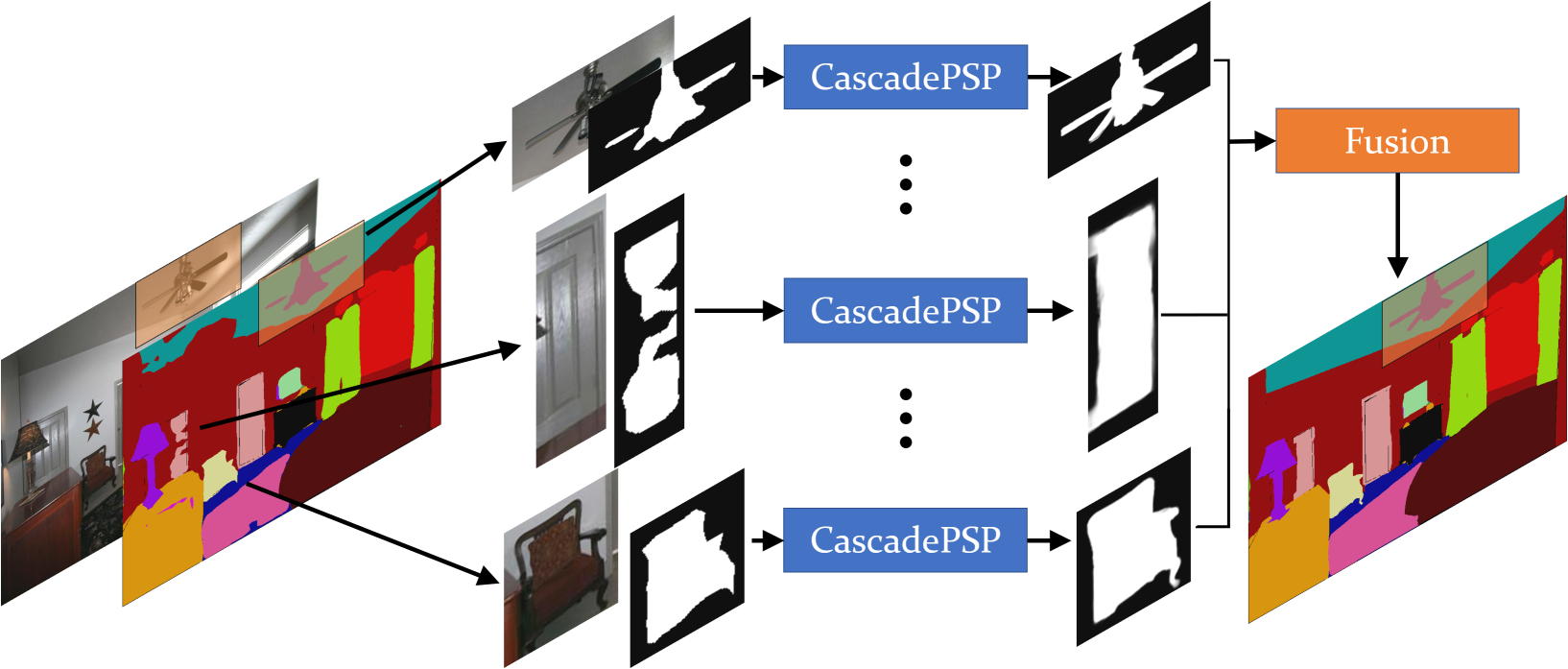}
\caption{Divide-and-conquer strategy in applying CascadePSP to scene parsing.}
\label{fig:scene_parsing}
\vspace{-0.3in}
\end{figure}

To extend CascadePSP to scene parsing, in the presence of dense classes where class competition may be problematic, we propose a divide-and-conquer approach to independently refine each semantic object using our pretrained network, followed by integrating the results using a fusion function. Figure~\ref{fig:scene_parsing} overviews our strategy. 

We refine sufficiently large connected components for each semantic object independently by taking ROIs with 25\% padding. 
To handle overlapping regions, the na\"ive approach would be to use argmax on the output confidence which would lead to noisy results in regions where all the classes have low scores.
Instead, our fusion function is a modified argmax where if all the input class confidence have values lower than $0.5$, we fall back to the original segmentation. 

\begin{figure}[h]
\centering
\minipage[t]{0.95\linewidth}
\centering
	\includegraphics[width=\linewidth]{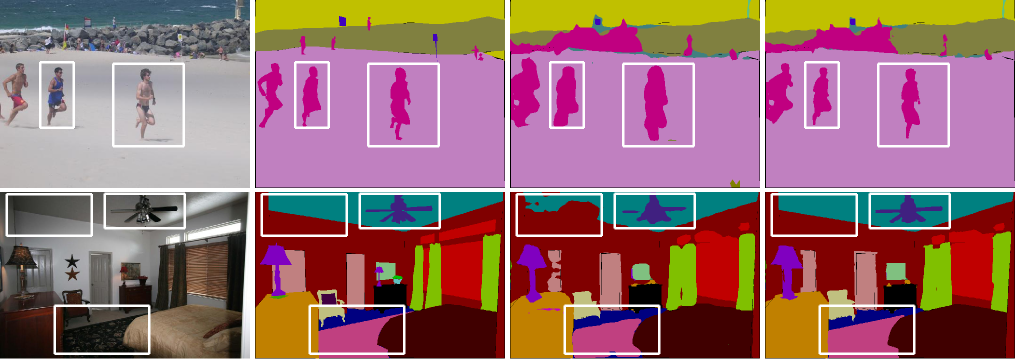}
	\includegraphics[width=\linewidth]{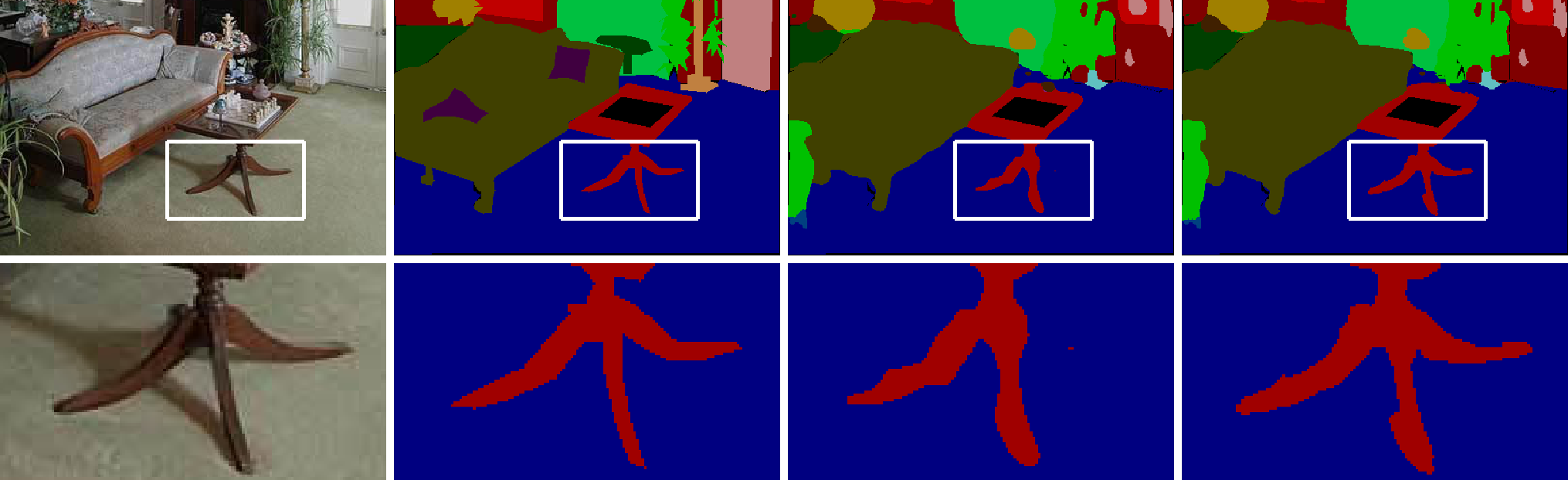}
	\minipage[t]{0.24\linewidth}
    	\begin{center}
        	\textbf{Image}
    	\end{center}
	\endminipage\hfill
	\minipage[t]{0.24\linewidth}
    	\begin{center}
        	\textbf{GT}
    	\end{center}
	\endminipage\hfill
	\minipage[t]{0.24\linewidth}
    	\begin{center}
        	\textbf{Input}
    	\end{center}
	\endminipage\hfill
	\minipage[t]{0.24\linewidth}
    	\begin{center}
        	\textbf{Ours}
    	\end{center}
	\endminipage\hfill
\endminipage
\vspace{0.05in}
\caption{Refinement results in the ADE20K validation set. 
    Top two rows: PSPNet. Bottom two rows: RefineNet.}
\label{fig:ade_fine_results}
\vspace{-0.25in}
\end{figure}

Here, we evaluate our model on the validation set of ADE20K~\cite{zhou2017scene}. 
As the ADE20K dataset contains ``stuff'' background classes (see supplementary material) that are not strong in objectness and too different from our training data, we have attenuated their output scores to focus on foreground refinement.
Note that refining the foreground objects can still help with background refinement since the argmax operation takes both confidence scores into consideration.
Table~\ref{tab:multiclass_table} tabulates the results which show that our model produces higher quality segmentation. Figure~\ref{fig:ade_fine_results} shows sample qualitative evaluation.

\begin{table}[t]
	\centering
	\begin{tabular}{l|c|c}
		Methods & mIoU (\%) & mBA (\%) \\
		\hhline{|=|=|=|} 
		\multicolumn{3}{c}{\textbf{ADE20K}} \\
		\Xhline{2\arrayrulewidth}
		RefineNet~\cite{lin2017refinenet} & 41.47 & 55.60 \\
		\quad (+) Ours & $\mathbf{42.20_{\uparrow0.73}}$ & $\mathbf{56.67_{\uparrow1.07}}$ \\
		\hline
		EncNet~\cite{zhang2018context} & 42.20 & 55.29 \\
		\quad (+) Ours & $\mathbf{43.19_{\uparrow0.99}}$ & $\mathbf{57.29_{\uparrow2.00}}$ \\
		\hline
		PSPNet~\cite{zhao2017pyramid} & 43.10  & 57.03  \\
		\quad (+) Ours & $\mathbf{43.83_{\uparrow0.73}}$ & $\mathbf{58.13_{\uparrow1.10}}$ \\
	\end{tabular}
    \caption{Comparison between different methods with and without our refinement on the ADE20K validation set.}
	\label{tab:multiclass_table}
	\vspace{-0.2in}
\end{table}

\section{Conclusion}
\vspace{-0.05in}
We propose CascadePSP, a general segmentation refinement framework for refining any input segmentations and achieve a higher accuracy without any finetuning afterward. CascadePSP performs high-resolution (up to 4K) segmentation refinement even our model has \textit{never} seen any high-resolution training images. 
With a single refinement module trained on low-resolution data without any finetuning, the proposed Global step refines the entire image and provides sufficient image context for the subsequent Local step to perform full-resolution high-quality refinement.
We hope this work can contribute to more high-resolution computer vision tasks in the future.
\vspace{-1em}

{\small
\paragraph{Acknowledgements}
We thank Gary Jing Yang Zhang for fruitful discussion during his exchange semester at HKUST.
}

\contourlength{1.5pt}

{\small
\bibliographystyle{ieee_fullname}
\bibliography{ms}
}

\end{document}